\documentclass[letterpaper, 10 pt, conference]{ieeeconf}
\IEEEoverridecommandlockouts 
\usepackage{amsmath,amsfonts}
\usepackage{array}
\usepackage{textcomp}
\usepackage{stfloats}
\usepackage{url}
\usepackage{verbatim}
\usepackage{graphicx}
\usepackage{cite}
\usepackage{amssymb}
\usepackage{soul}
\usepackage{pdflscape}
\usepackage{rotating}
\usepackage{colortbl}
\usepackage{pdfpages}
\usepackage{multirow}
\usepackage{xcolor}
\sethlcolor{yellow}
\usepackage{array}
\usepackage[linesnumbered,ruled]{algorithm2e} 
\usepackage{setspace}
\usepackage{algorithmic}
\RestyleAlgo{ruled}
\usepackage{threeparttable}
\usepackage{hyperref}
\hypersetup{
    colorlinks=true,
    linkcolor=blue,filecolor=magenta,      
    urlcolor=magenta}

\SetCommentSty{mycommfont}
\title{\LARGE \bf
CoBT: Collaborative Programming of Behaviour Trees from One Demonstration for Robot Manipulation
}
\author{Aayush Jain$^{1,2}$,  Philip Long$^{3}$, Valeria Villani$^{4}$, John D. Kelleher$^{5}$, Maria Chiara Leva$^{1}$% <-this % stops a space
\thanks{This work is supported by the CISC project which is funded by the European Union’s Horizon 2020 MSCA-ITN grant agreement no. 955901; and ADAPT Research Centre which is funded by SFI Research Centres Programme and is co-funded under the European Regional Development Fund (ERDF) through Grant 13/RC/2106\_P2.}% <-this % stops a space
\thanks{$^{1}$Technological University Dublin; $^{2}$Irish Manufacturing Research; $^{3}$Atlantic Technological University; $^{4}$ARS Control Lab, University of Modena and Reggio Emilia; $^{5}$ADAPT Research Centre, School of Computer Science and Statistics, Trinity College Dublin. Email: {\tt\small aayush.jain@imr.ie}}%
}

\begin{document}
\setlength\textfloatsep{3pt}
\setlength\floatsep{3pt}
\setlength\dbltextfloatsep{10pt}

\maketitle
\thispagestyle{empty}
\pagestyle{empty}

\begin{abstract}
Mass customization and shorter manufacturing cycles are becoming more important among small and medium-sized companies. However, classical industrial robots struggle to cope with product variation and dynamic environments. In this paper, we present CoBT, a collaborative programming by demonstration framework for generating reactive and modular behavior trees. CoBT relies on a single demonstration and a combination of data-driven machine learning methods with logic-based declarative learning to learn a task, thus eliminating the need for programming expertise or long development times. The proposed framework is experimentally validated on 7 manipulation tasks and we show that CoBT achieves $\approx$ 93\% success rate overall with an average of 7.5s programming time. We conduct a pilot study with non-expert users to provide feedback regarding the usability of CoBT. 
More videos and generated behavior trees are available at: \href{https://github.com/jainaayush2006/CoBT.git}{https://github.com/jainaayush2006/CoBT.git}.
\end{abstract}

% \begin{IEEEkeywords}
% Interactive Task Learning, Programming from Demonstration, Behavior Tree,  Online Failure Detection
% \end{IEEEkeywords}

\section{Introduction} \label{sec:introduction}

Intelligent robotic systems can improve factory ergonomics and productivity by collaborating with humans. Furthermore, as industry shifts toward mass customization, reprogramming robots using traditional methods will result in higher downtime and increased cost. Thus, there is a need for agile methods that can generate reactive and modular robot programs, for instance, programming by demonstration (PbD) which collaboratively generates task models using human assistance.

PbD (Learning from Demonstration or Imitation Learning) is a promising approach to simplify robot programming to advance human-centric automation in Industry 5.0 \cite{jain2022evaluating}. PbD enables human collaborators to transfer skills to a robot through task demonstration, though  developing such systems to handle the high variability and uncertainty in real-world environments remains a challenge~\cite{laird2017interactive}. Most existing PbD methods generalize \textit{motion-level} (low-level learning) plans \cite{kyrarini2019robot, lentini2020robot, knaust2021guided} which are often insufficient to accomplish complex tasks where \textit{task-level} (high-level learning) knowledge \cite{gustavsson2022combining,9844164, eiband2023collaborative, kroemer2015towards} is also required. However, learning on both levels is computationally expensive and time-consuming, as it requires multiple demonstrations or self-exploration to generalise. 

\begin{figure}[!t]
\centering
\includegraphics[width=\columnwidth]{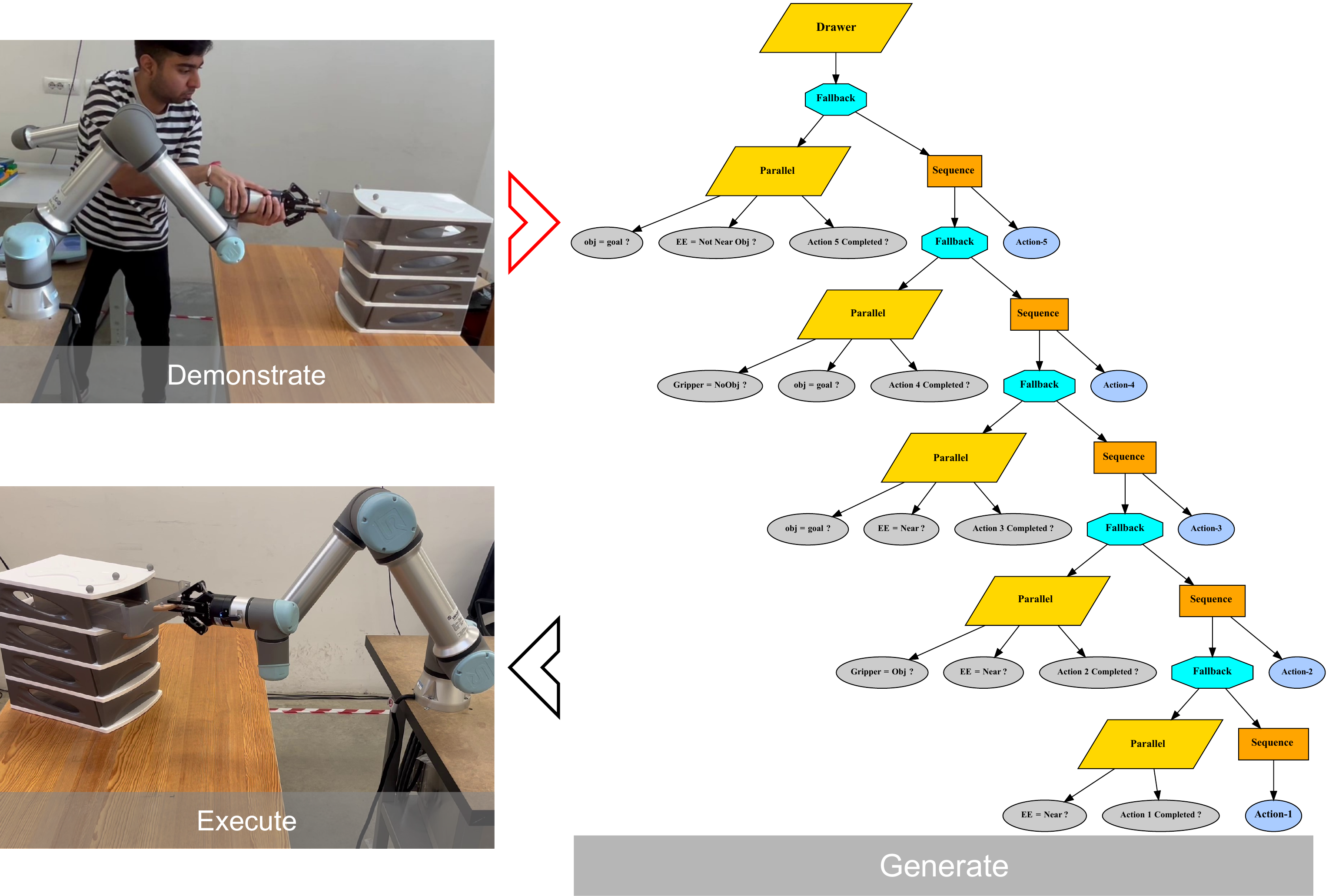}
\caption{CoBT: a collaborative programming framework that generates deployable behavior trees from one demonstration. This example shows the work-flow of programming opening a drawer task.}
\label{fig:Overview}
\end{figure}

To this end, a novel \textbf{Co}llaborative programming of \textbf{B}ehavior \textbf{T}rees (CoBT) framework is proposed. A single demonstration is segmented into symbolic states (Algorithm \ref{alg:segmentation}) and primitive actions (Algorithm \ref{alg:Primitive}) to automatically generate a BT (Algorithm \ref{alg:BTgen}). The generated BT provides a high-level control structure that enables the robot to react to environmental changes, while primitive actions in the form of Dynamic Movement Primitives (DMPs) enable the robot to generalize its movements to different environments. The learned BTs are adapted to user-defined goal to accomplish complex task multi-level tasks. Our key contributions are: (1) a collaborative programming framework that generates deployable BTs from one demonstration, (2) an automatic generation of state-action pairs to form atomic BTs which can perform complex motions and (3) a Logic-based Declarative Learning algorithm (Algorithm \ref{alg:BTgen}) capable of resolving task and motion constraints and generating BTs by composing action primitives. The paper is organized as follows: in Section \ref{sec:related_work}, the related work is presented; the CoBT framework and its components are introduced in Section \ref{sec:ITL}. The results of experiments across 7 tasks and pilot study conducted are discussed in Section \ref{sec:experiments}. Lastly, in Section \ref{sec:conclusion}, we conclude the discussion with limitations and future work.

\section{Related Work} \label{sec:related_work}
In this section, frameworks that either program from one demonstration or generate BTs from multiple demonstration are outlined, as shown in Table \ref{table:comparison}. We compare the frameworks based on:  number of demonstrations, segmentation method (manual or automatic), motion learning (movement primitives or point-to-point), reactivity, and modularity.

To improve task generalization, a demonstration must be segmented into a sequence of sub-tasks or primitive actions \cite{osa2018algorithmic}. Segmentation can be automatic \cite{diehl2021automated} and  \cite{scherf2023interactively}  or \cite{zanchettin2023symbolic, liang2022iropro, lentini2020robot, french2019learning, gustavsson2022combining, knaust2021guided} done manually. While \cite{diehl2021automated} and \cite{scherf2023interactively} utilize pre-defined rules and pre-trained classifiers for segmentation, CoBT can automatically segment even unseen actions. We achieve this by employing a parameter-less velocity-based change-point detection \cite{truong2020selective} algorithm and grounding state variables. Compared to other methods, this simplifies user task demonstrations and reduces overall complexity.

Among the frameworks described in Table \ref{table:comparison}, \cite{zanchettin2023symbolic, diehl2021automated, liang2022iropro, gustavsson2022combining, french2019learning, scherf2023interactively}, the main approach for motion generation is point-to-point (P2P), focusing on learning keyframes and way-points. However, this approach fails to preserve the shape of trajectories, limiting its functionality to executing variations of pick-and-place tasks.
By contrast, Lentini et al. \cite{lentini2020robot} proposed learning DMPs \cite{ude2014orientation} from one simple demonstration without prior constraints. However, their method lacks reactivity to environmental changes due to an absence of state-action mapping. To achieve reactivity at both task and motion levels over a  variety of complex tasks, we learn DMP \cite{ude2014orientation} embedded atomic BT for each segment using the symbolic state variables. Moreover, we learn a set of DMPs for opposite trajectory to the one demonstrated, in order to generalise opposite start and end poses.

\begin{table}[!t]
\centering
\resizebox{0.9\columnwidth}{!}{
\begin{threeparttable}[b]
\small
\caption{Comparison with the related frameworks}
\label{table:comparison}
\begin{tabular}{c|c|c|c|c|c}
\hline
\textbf{Ref.} & \textbf{Demos.} & \textbf{Seg.} & \textbf{Motion} & \textbf{Reactive} & \textbf{Modular}\\
\hline
\cite{lentini2020robot} & One & - & DMP & No & No\\
\cite{zanchettin2023symbolic} & One & Manual & P2P & No & No\\
\cite{diehl2021automated} & One  & Auto\tnote{*} & P2P & No & Yes\\
\cite{liang2022iropro} & One & Manual & P2P & No & Yes\\
\cite{knaust2021guided}& Multi. & Manual & P2P/ProMP & Yes & Yes\\
\cite{gustavsson2022combining} & Multi. & Manual & P2P & Yes & No\\ 
\cite{french2019learning} & Multi. & Manual & P2P & Yes & No \\ 
\cite{scherf2023interactively} & Multi. & Auto\tnote{*} & P2P & Yes & No\\
\textbf{CoBT} & \textbf{One} & \textbf{Auto} & \textbf{DMP} & \textbf{Yes} & \textbf{Yes}\\
\hline
\end{tabular}
\begin{tablenotes}
    \item [*]Require pre-training or domain knowledge
\end{tablenotes}
\end{threeparttable}
}
\end{table} 

In PbD, data-based learning techniques require a large amount of data to generalize \cite{sunderhauf2018limits}. \cite{lentini2020robot} proposed a data-efficient methodology that combines data-based learning and symbolic AI. Therefore, we explore logic-based declarative learning \cite{kordjamshidi2022declarative} to generate BTs. Declarative learning represents robot manipulation knowledge in the form of logical rules which can be used to generate explainable and transparent policies from the data through logical reasoning processes. The existing techniques to build BTs have limitations since they either generate BTs at runtime \cite{9844164, french2019learning} or are trained in a simulator\cite{gustavsson2022combining, scherf2023interactively}, making their outcome non-deterministic and limiting their scalability in actual industrial settings.

In general compared to Finite State Machines (FSM) \cite{grollman2010incremental}, Hierarchical Task Networks (HTN) \cite{nejati2006learning} and Hidden Markov Models (HMM) \cite{kroemer2015towards}, Behaviour Trees (BT) \cite{colledanchise2018behavior} have proven to be a more reactive and modular approach to robot control. They allow the creation of complex behaviors by combining atomic BTs \cite{colledanchise2019towards} i.e., simple sub-behaviors, in a hierarchical structure. Further, BTs are a human-readable and explainable control architecture. To the best of our knowledge, BTs have been used as high-level task plans for PbD only in \cite{french2019learning, gustavsson2022combining, scherf2023interactively, 9844164}. \cite{french2019learning} learn a decision tree (DT) from demonstration, which is converted into a BT. \cite{gustavsson2022combining} and \cite{scherf2023interactively} learn task constraints from demonstration and combine them with backchaining \cite{colledanchise2019towards} to generate BTs. \cite{9844164} proposed a method to learn BTs from natural language instructions by creating a mapping between logical forms and action BTs. However, these techniques require multiple demonstrations and generate rigid BTs that lack modularity. Conversely, CoBT requires one demonstration and the learned BT can be adapted to achieve a new goal or can be chained with other BTs to perform long-horizon tasks.

Valassakis et al. \cite{valassakis2022demonstrate} presented an approach to one-shot PbD, DOME, where visual servoing is used to reach the user-specified pose and then the demonstrated velocities are replayed to imitate. Despite DOME's promising results, the generated policies are highly dependent on user-specified pose and are limited to short-horizon tasks like only pick or insert. CoBT is the only framework capable of generating deployable, reactive, and modular robot programs in the form of BTs through one demonstration which are capable of executing complex motions and long-horizon tasks.

\section{Methodology} \label{sec:ITL}
\begin{figure}
\begin{center}
    \includegraphics[width=\columnwidth]{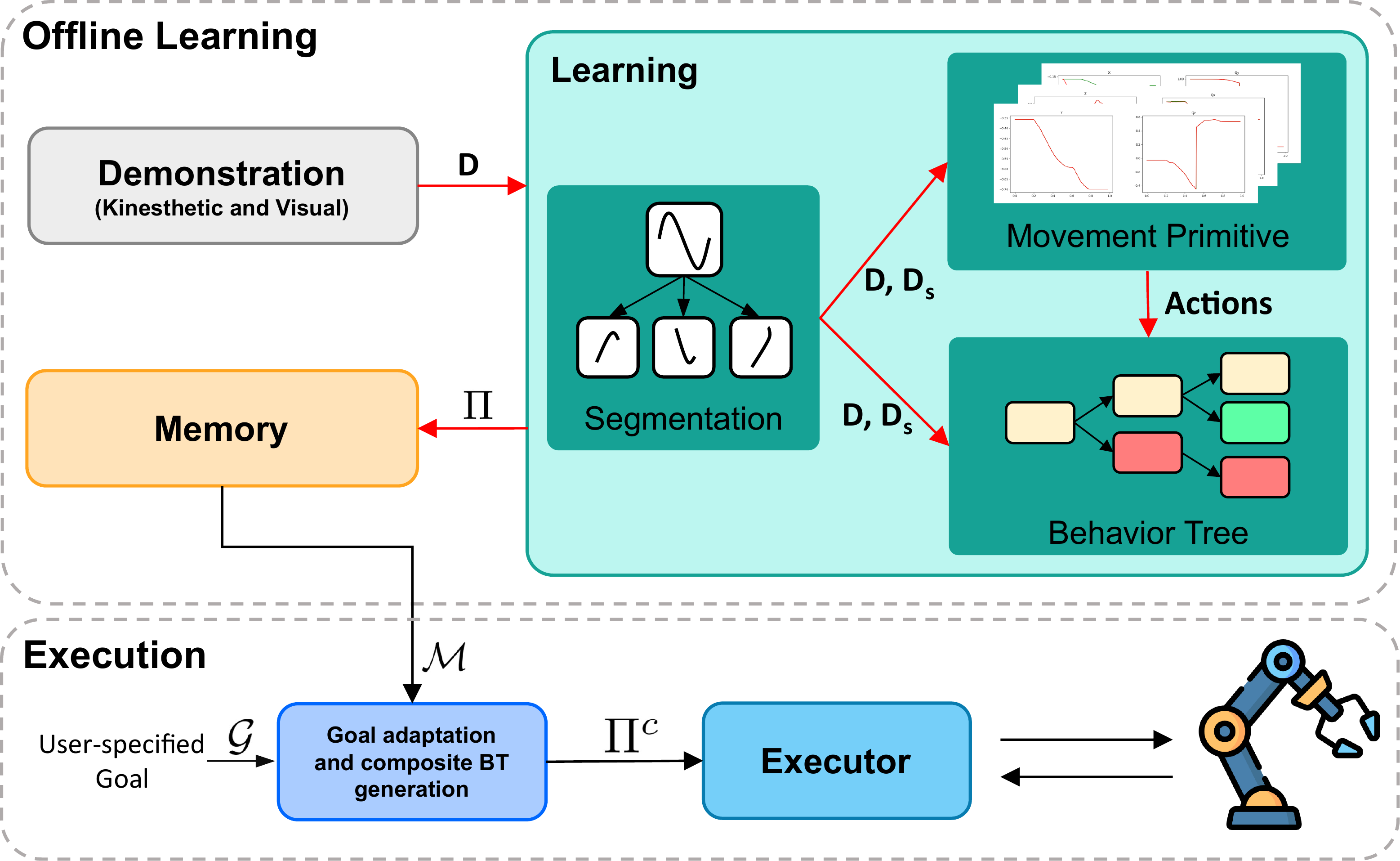}
     \caption{A high-level workflow of offline learning and execution in CoBT. The red lines, depict the task (BT) and action (DMP) policies that are learned offline through PbD. During execution, depicted with black lines, the user-defined goal is abstracted to adapt the BT parameters. The generated BT decides \textit{``what"} action to execute, and DMPs control \textit{``how"} the action should be executed based on the current environmental conditions.}\label{fig:framework}
\end{center}
\end{figure}

CoBT, shown in Fig. \ref{fig:framework}, comprises- \textit{demonstration, learning, memory, goal adaptation and composite BT generation, and executor} and is organized into: \textit{offline learning}, and \textit{execution}. 
During \textit{offline learning}, the user demonstrates the task, using the \textit{demonstration module}, which records them as multivariate time series data. This data subsequently serves as an input for the \textit{learning module}, which performs automatic segmentation (Algorithm \ref{alg:segmentation}), primitive learning (Algorithm \ref{alg:Primitive}) and generates BTs (Algorithm \ref{alg:BTgen}).

Algorithm \ref{alg:segmentation} automatically decomposes the demonstration into different phases. State variables, shown in Table \ref{table:grounding}, are grounded depending on real-world sensory data at each segment. As a result, the demonstration is segmented into a set of state-action pairs. Each action is then encoded using DMP (Algorithm \ref{alg:Primitive}) and the action sequence is formalized in a BT (Algorithm \ref{alg:BTgen}). Actions are embedded in a BT's leaf node and the state variables are used as task and action constraints. 
\textit{Task constraints} refer to the relative sequencing of primitive actions and the fulfillment of goal conditions, while \textit{action constraints} are the pre-conditions for action execution and the effects that must be met to consider the action completed.
During the \textit{execution phase}, a BT is generated to accomplish the user-specified goal using the \textit{memory} from the learned task (Algorithm \ref{alg:Goal_abs}). The high-level objective of ``what action to  perform?" and the low-level objective of ``how to perform the action?" is adapted through BT and DMPs respectively based on the environmental state, as follows.

\begin{table}[!t]
\resizebox{\columnwidth}{!}{
\begin{threeparttable}[b]
\small
\caption{State variables and respective grounding
}
\label{table:grounding}
\begin{tabular}{c|p{5.7cm}}
\hline
\textbf{State variables} & \textbf{Grounding conditions}\\
\hline
${g}_i$ & A gripper has opened ({\tt Open}) or closed ({\tt Close}) its fingers\\
${o}_i$ & Target object nearer to another object ({\tt Near}) compared to $i-1^{th}$ segment or object on goal position ({\tt On\_goal})\\
${e}_i$ & End-effector {\tt Near} or {\tt !Near} the object\\
\hline
\end{tabular}
\end{threeparttable}
}
\end{table}
% - Alg 1:
%   - Not clear how N relates to D. Is this |D|? Or user
% specified (but if it is, why isn't it an input)
%   - While loop with no increment; Maybe "for t in D" or
% similar could be more precise
%   - B (number of segments?) not defined in the algorithm
\begin{algorithm}[!t]
\caption{Segmentation}\label{alg:segmentation}
\KwIn{Demonstration dataset $\textbf{D}$}
\KwOut{Segmented dataset $\textbf{D}_s$}
\While{$t \leq N$}{
    $\textbf{v}_N \gets v_t$ \tcp*{$v_t$ = $({v_{x_t}}^2$ + ${v_{y_t}}^2$ + ${v_{z_t}}^2)^{\frac{1}{2}}$}
}
$\textbf{v}_N = [v_1, v_2, ..., v_N]$\\
$\textbf{i}$ = $\textit{change\_point\_detection}(\textbf{v}_N) \in \mathbb{R}^{1 \times B}$\\
$\textbf{g}_i$ = $[g_{i_1}, g_{i_2}, ...]$; $\textbf{o}_i$ = $[o_{i_1}, o_{i_2}, ...]$; $\textbf{e}_i$ = $[e_{i_1}, e_{i_2}, ...]$\\
$\textbf{States}$ = $[\textbf{g}_i^T, \textbf{o}_i^T, \textbf{e}_i^T] \in \mathbb{R}^{B \times 3}$\\
\KwRet $\textbf{D}_s$ = $[\textbf{i}^T, \textbf{States}] \in \mathbb{R}^{B \times 2}$
% \textbf{where}, $D_s$ \in \mathbb{R}^{x \times 4}\\
\end{algorithm}

\subsection{Segmentation}\label{subsec:segmentation}
We assume that an action's start and end will alter the end-effector's velocity which is then used for segmentation in Algorithm \ref{alg:segmentation}. An offline change point detection algorithm \cite{truong2020selective} with penalty function obtains segments $\textbf{i} = [ i_1 \ldots i_B ]$, based on the velocity $\textbf{v}_N = [ v_1 \ldots v_t \ldots v_N ]$, where $B$ is the total segment number and $v_t$ is the end effector's translational velocity norm at timestamp $t$. Lastly, to generate the BTs, the state variables at each segment are grounded. To avoid false segmentation, the segments where the end-effector velocity or symbolic states are unchanged are ignored. A combination of kinesthetic and visual teaching interfaces is used to record demonstrations in the form of multivariate time series. For $N$ timestamps, the end effector pose $\textbf{x}^{e}_t = [x,y,z,w,qx,qy,qz]^T$, gripper state $g_t$ and object poses $\textbf{x}^{o}_t = [x,y,z,w,qx,qy,qz]^T$ are recorded to create a dataset $\textbf{D}$ where $t = 1,...,N$:
\begin{equation}
\begin{cases}
\textbf{X}^{e} \in \mathbb{R}^{7 \times N},\textbf{X}^{e}=[\textbf{x}^{e}_0,...,\textbf{x}^{e}_N],\\
\textbf{g} \in \mathbb{R}^{1 \times N},\textbf{g}=[g_0,...,g_N],\\
\textbf{X}^{o} \in \mathbb{R}^{7 \times N},\textbf{X}^{o}=[\textbf{x}^{o}_0,...,\textbf{x}^{o}_N].\\
\end{cases}
\end{equation}
% }
Three symbolic state variables, the gripper state $\textbf{g}_i$, the target object state $\textbf{o}_i$, and the end-effector state $\textbf{e}_i$ at the $i^{th}$ segment, are necessary to generate a BT. In the final step, the values in $\textbf{D}$ are grounded into symbolic states variables for each segment according to conditions from Table \ref{table:grounding}. The grounded state variables are stored in the vector $\textbf{States}$, which is in turn stored in a segmented dataset $\textbf{D}_s$ with segments $\textbf{i}$. A 5 cm threshold is used to ground $\textbf{o}_i$ and $\textbf{e}_i$.

\subsection{Primitive Learning}\label{subsec:Primitive}

The low-level motion policy is learned using DMPs \cite{ude2014orientation} as they require only one demonstration to generalize and can adapt to different start and end poses. Algorithm \ref{alg:Primitive} shows that for each $Action^b$ we train a set of DMPs, $\tau^b$, for $\textbf{X}^e$ and $\textbf{g}$, compute the relative object $\mathcal{O}^b$ and a transformation matrix $^{\mathcal{O}}\textbf{T}_e$ for the end-effector at the segment end. Relative object $\mathcal{O}^b$ is defined as the nearest object to the end-effector w.r.t. which end-effectors distance decreased between the action. 

% at the end of the action and \hl{if the end-effector is closed then the second nearest object.}
% \hl{ JK note on highlighted text: what happens if we close the end effector in order to grasp the target object but we missed, then should we not still be looking at the closest object even if the end effector is closed?} 

\subsection{Automatic behavior tree generation}\label{subsec:BTgen}
\begin{figure}[t]
\centering
\includegraphics[width=\columnwidth]{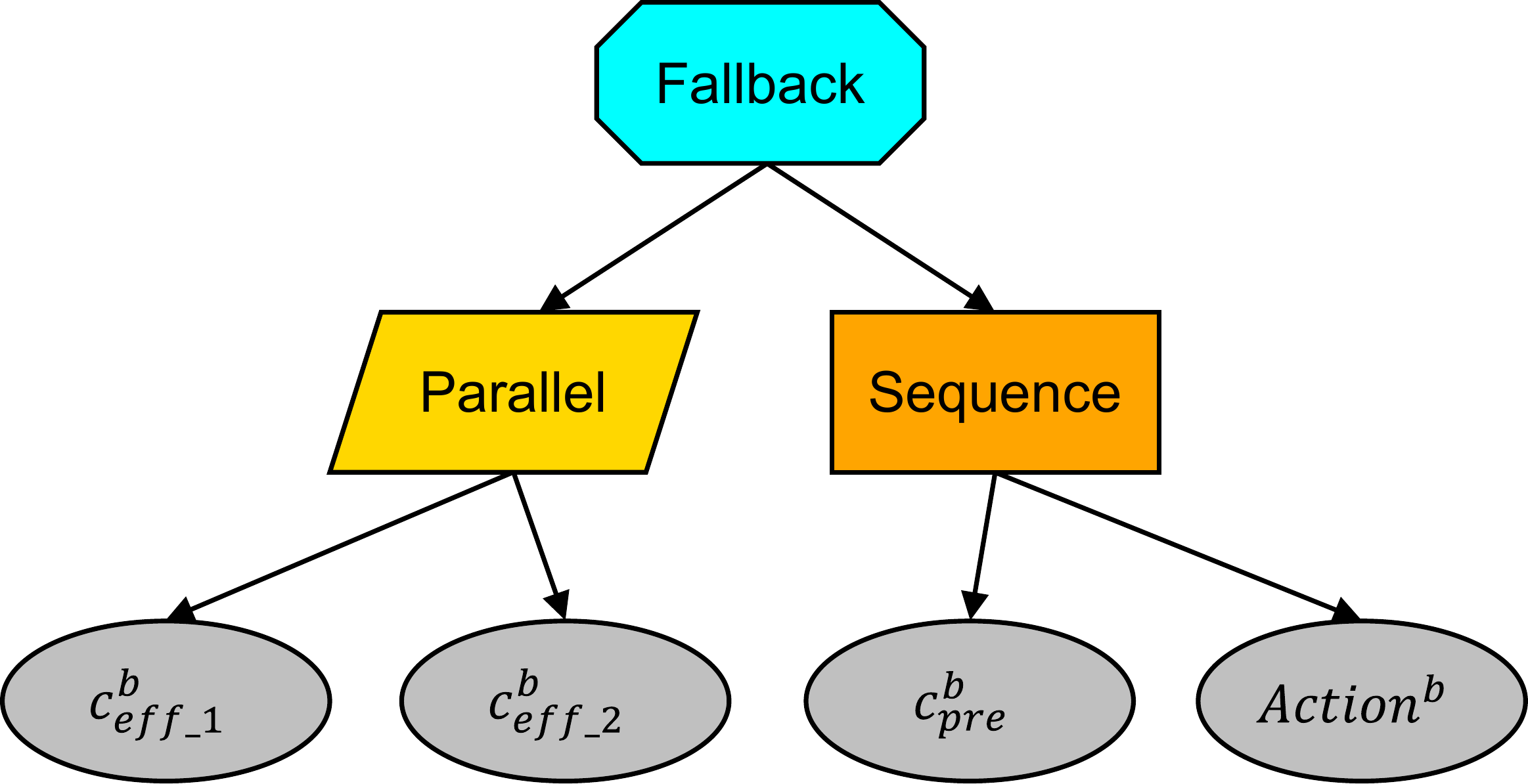}
\caption{Atomic BT. The effect-conditions $c^{i}_{eff}$ are achieved by action $Action^i$. To execute the action, pre-condition $c^i_{pre}$ should satisfy, The BT starts with a fallback node connected to a parallel node for effect-conditions and a sequence node for action and pre-conditions.}
\label{fig:atomicBT}
\end{figure}

% - Alg 2:
%   - Would be good to link X, g etc to the input, which is
% the segmented data or original data (since you pass in
% pother)
\begin{algorithm}[!t]
\caption{Primitive Learning}\label{alg:Primitive}
\KwIn{Demo. dataset $\textbf{D}$, Segmented dataset $\textbf{D}_s$}
 \KwOut{Primitive Actions $\textbf{Actions} \in \mathbb{R}^{B-1 \times 3}$}
% $\textbf{Actions} \gets \emptyset$\\
\While{$b < B$}{
    $\tau^b = trainDMP(\textbf{X}^e[i^b:i^{b+1}], \textbf{g}[i^b:i^{b+1}])$\\
    $\mathcal{O}^b = Relative\_Object(i^{b+1})$\\
    $\textbf{T}^b = EE\_transformation(\mathcal{O}^b, i^{b+1})$\\
    $\textbf{Actions} \gets [\tau^b, \mathcal{O}^b, \textbf{T}^b]$
}
\end{algorithm}

% - Alg 3
%   - Line 14: when b = B (as per condition) can States^b+1
% go out of range? Double check loop condition
\begin{algorithm}[!t]
\SetKwProg{Fn}{Function}{:}{}
\SetKwFunction{Cond}{CondAbstraction}
\SetKwFunction{BT}{CondToTree}
% \setstretch{1.2}
\caption{Behavior Tree Generation
% \hl{B  isn't initialized anywhere should it be?}
}\label{alg:BTgen}
% \KwIn{Demonstration dataset $D\in\{P,G,O\}$, Segmented dataset $D_s \in \{{i}, States\}$}
\KwIn{Demo. dataset $\textbf{D}$, Segmented dataset~$\textbf{D}_s$, 
Primitive Actions $\textbf{Actions}$}
\KwOut{Behavior Tree $\Pi$}
$\Pi \gets \emptyset$\\
\Begin(\textit{Behavior Tree Generation}){
    \textit{$\mathcal{T}$} = \Cond{\textbf{States, Actions}}\\
    $\Pi$ = \BT{$\mathcal{T}$}\\
    $\mathcal{M} \gets \Pi$\\
    % \KwRet $\mathcal{M}$
  }
\Fn{\Cond{$\textbf{States}, \textbf{Actions}$}}{
    $\mathcal{T} \gets \emptyset$, $b \gets 2$\\
    \While{$b\leq B$}{
        $c_{pre}^b \gets \emptyset$, 
        $c_{eff}^b \gets \emptyset$\\
        \While{$n<3$}{
            \If{$States_{n}^b \neq States_{n}^{b-1}$}{
                $c_{eff}^b \gets States_{n}^b$\\
                \If{$States_{n}^b=States_{n}^{b+1}$}{
                    $c_{pre}^{b+1} \gets States_{i}^b$\\}
            }
            $\mathcal{T} \gets Action^b, c_{pre}^{b+1}, c_{eff}^b$
        }
    }
    \KwRet $\mathcal{T} \in \mathbb{R}^{B-1} $\;
  }

\Fn{\BT{$\mathcal{T}$}}{
    $b \gets 1$\\
    \While{$b \leq B-1$}{
        $Parallel \gets \textbf{c}^b$\\
        \eIf{$\Pi = \emptyset$}{
            $Sequence \gets Action^b$
        }{
            $Sequence \gets \Pi,Action^b$
        }
        $Fallback\gets Parallel, Sequence$\\
        $\Pi \gets Fallback$\\
    }
    \KwRet $\Pi$\\
  }
\end{algorithm}

The proposed method in Algorithm \ref{alg:BTgen} is inspired by logic-based declarative learning \cite{kordjamshidi2022declarative} and atomic BT \cite{colledanchise2019towards}. Given a fully-observable environment, the task is simplified by simply identifying the state transitions during the demonstration \cite{jimenez2012review}. If the deterministic effects in the environment can be observed, then the state transitions will happen only when an action is performed. As the demonstration is segmented into its lowest form, a state transition occurs due to a single action. Hence, the states at the segment start are considered action pre-conditions, denoted $c_{pre} \in \{{g}_i, {o}_i, {e}_i\}$, while the altered states at the segment end are denoted the effects, $c_{eff} \{{g}_i, {o}_i, {e}_i\}$. However, conditions that are pre-conditions for the next action may be ignored as they are unchanged during a particular segment. Thus, the conditions that are pre-conditions \textit{for the next action} are also added as action constraints (line 14 of Algorithm \ref{alg:BTgen}), using the {\tt CondAbstraction} function which returns a tuple $\mathcal{T}$ in the form of $Action^b$ sequence with the corresponding constraint vector $\textbf{c}^b \in \{c_{pre}^{b+1}, c_{eff}^b\}$, i.e.,
\begin{equation}
\mathcal{T}  = \{(Action^1, \textbf{c}^1),..., (Action^{B-1},\textbf{c}^{B-1})\}.\\
\end{equation} 
Each action, with its state conditions, can be converted into an atomic BT~\cite{colledanchise2019towards}, which aims to satisfy the effect conditions and invokes the action if any of the effect conditions are unmet and the preconditions satisfied. An illustrative atomic BT for an action with pre- and effect-condition is shown in Fig. \ref{fig:atomicBT}. Through the {\tt CondToTree} function in Algorithm \ref{alg:BTgen}, actions of the $\mathcal{T}$ are converted into an atomic BT. Finally, each atomic BT is chained in a sequence assuming that the pre-condition for an action to occur is the previous action's effects' success. The generated BT for a particular target and its respective $\textbf{Actions}$ and target object are saved in the memory dictionary $\mathcal{M}$.

\subsection{Goal adaptation and composite BT generation}\label{subsec:Goal_abs}
% - Alg 4:
%   - Line 14 returns variables that don't get set in the alg

\begin{algorithm}[!t]
\SetKwProg{Fn}{Function}{:}{}
\SetKwFunction{Composite}{CompositeBT}
\SetKwFunction{GOAL}{AdaptGoal}
\caption{Goal adaptation and composite BT generation}\label{alg:Goal_abs}

\KwIn{Memory $\mathcal{M}$,
User-specified goal scene $\mathcal{G}$}
\KwOut{Composite BT $\Pi^c$}

\Begin{
\textit{found\_objects, new\_goal} = \GOAL{$\mathcal{G, M}$}

$\Pi^c$ = \Composite{found objects, new goal}
}

\Fn{\GOAL($\mathcal{G}, \mathcal{M})$}{
\textit{All\_objects = find\_all\_objects}($\mathcal{G}$)\\
\textit{found\_objects = match\_objects}($\mathcal{M},All\_objects$)\\
\If{$found\_objects \neq$ None}{
    \eIf{\textit{goal\_object\_found}}{\textit{save\_new\_goal}(w.r.t goal\_object)}{
    \textit{save\_new\_goal}(w.r.t. nearest\_non-goal\_object)}}
    \KwRet{found\_objects, new\_goal}}
    
\Fn{\Composite{found\_objects, new\_goal}}{
    \ForAll{found\_object}{$Sequence \gets \Pi(found\_object, new\_goal)$\\
    }
    $\Pi^c \gets Sequence$\\
    \KwRet{$\Pi^c$}}
\end{algorithm}

Real-world long-horizon tasks, such as placing an object in a drawer, can be hierarchically decomposed into a sequence of shorter-horizon sub-tasks, i.e., opening a drawer and placing an object. Thus, we exploit the modular nature of BTs to accomplish such tasks. The goal of the memorised BT is adapted and different adapted BTs are sequenced to form a composite BT. The aim is to show that the learned BTs can be combined for complex tasks without extra demonstrations, while emphasizing that hierarchical decomposition and long-horizon plan generation is not this work's primary goal.
Instead, off-the shelf high-level planning methods can be utilized to generate long-horizon plans, which can then be executed using the BTs developed in this study.

In this work, an effective goal abstraction method, Algorithm \ref{alg:Goal_abs}, is used to adapt and form composite BT. First, all the objects present in the user-specified goal scene $\mathcal{G}$ are matched with the objects in memory $\mathcal{M}$. If an object is found, the new goal position with respect to the goal object or nearest non-memorized object is saved which will be treated as the new goal. 
When multiple objects are matched, then the object with same goal object in the memorised BT is given higher preference in the execution sequence.

\section{Experiments and Results} \label{sec:experiments}
\subsection{Experimental setup}\label{subsec:set-up}
To validate CoBT, a UR5e collaborative robot equipped with a Robotiq 2-finger adaptive gripper is used, with OptiTrack motion capture system used for object detection. The system is implemented with ROS Noetic and ROS Foxy, with change point detection using the  \textit{ruptures}\footnotemark{} Python library \cite{truong2020selective}\footnotetext{\href{https://github.com/deepcharles/ruptures}{https://github.com/deepcharles/ruptures}}, DMPs using the \textit{movement\_primitives}\footnotemark{} Python library\footnotetext{\href{https://github.com/dfki-ric/movement\_primitives}{https://github.com/dfki-ric/movement\_primitives}} and BTs with the \textit{py\_trees}\footnotemark{}\footnotetext{\href{https://github.com/splintered-reality/py\_trees}{https://github.com/splintered-reality/py\_trees}} and \textit{py\_trees\_ros}\footnotemark{}\footnotetext{\href{https://github.com/splintered-reality/py\_trees\_ros}{https://github.com/splintered-reality/py\_trees\_ros}} Python libraries. The \textit{py\_trees} interface acts both as an operator display for task progress and carries out decision-making. 

\subsection{Evaluation Tasks} \label{subsec:task}
\begin{figure}[t]
\centering
\includegraphics[width=\columnwidth]{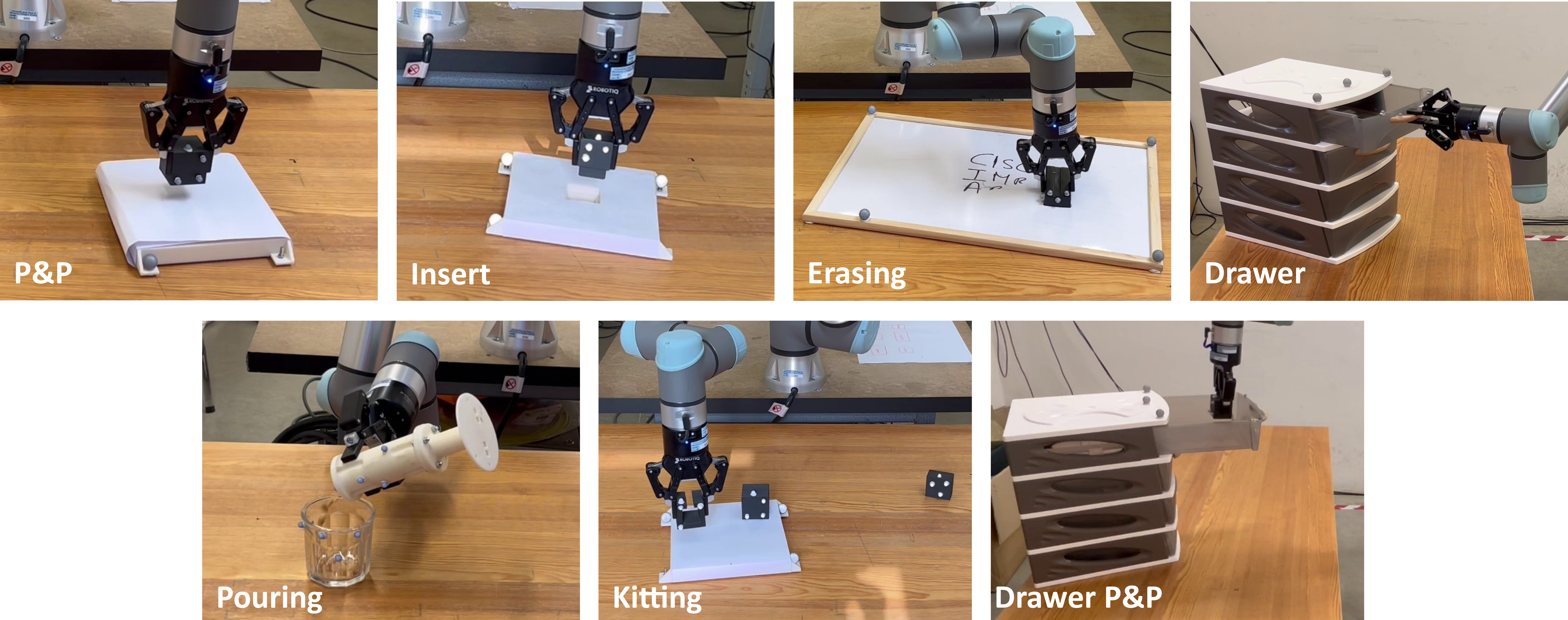}
\caption{7 evaluation tasks that include mix of complex and P2P trajectory executions, and short and long-horizon tasks with multi-level goals.}
\label{fig:tasks}
\end{figure}

For evaluation, 7 tasks, shown in Fig. \ref{fig:tasks}, are chosen. A \textit{pick-and-place} (\textit{P\&P}) task where a cube is picked and placed on the white tray. An \textit{insert} task where a \textit{35mm}x\textit{35mm} cube is inserted inside a slot of \textit{40mm}x\textit{40mm}. An \textit{erasing} task where a white board is erased using a duster. A \textit{drawer} task where the robot must open a drawer. A \textit{pouring} task where the contents of one cup are transferred to the other cup. A \textit{kitting} task where a user-specified kit is prepared using three objects. A \textit{drawer P\&P} task where a robot opens a drawer and places a cube inside. No task specific alterations in the algorithms are made to validate the generalizability of CoBT. 

\subsection{Demonstration to Behavior Tree} \label{subsec:demo_to_BT}

\begin{figure}[!t]
\centering
\includegraphics[width=\columnwidth]{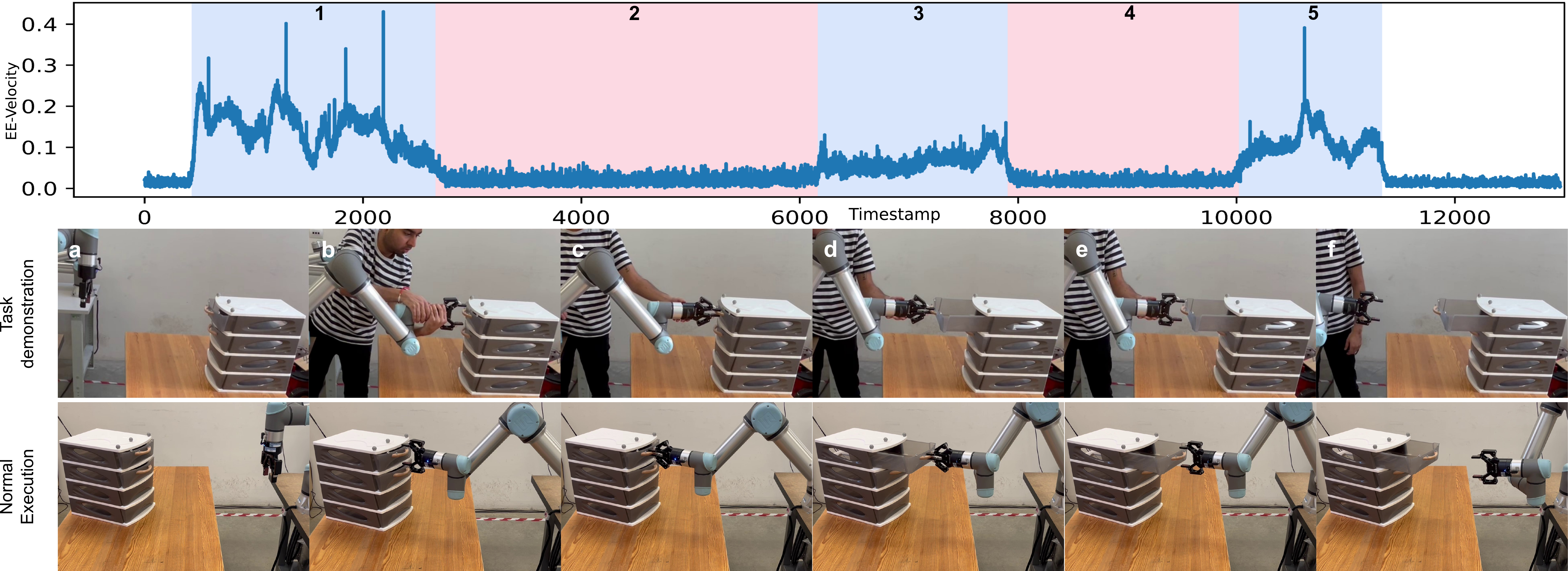}
\caption{(Top) Segmentation based on velocity and gripper state. (Middle) Transitions during the \textit{drawer} task demonstration. For example, the state transition from a to b (middle) occurs due to action in segment 1 (top). (Bottom) a trial example of the generated policy under normal conditions.}
\label{fig:demo_timeline}
\end{figure}

This section describes the full pipeline for the \textit{drawer} task. Hand-guiding is used for demonstration as shown in Fig. \ref{fig:demo_timeline} (middle). The task is performed in one continuous demonstration during which relevant time-series data is recorded to create the dataset $\textbf{D}$ and in turn segment into actions as shown in Fig. \ref{fig:demo_timeline} (top). A state transition from a to b of Fig. \ref{fig:demo_timeline} (middle) occurs due to $Action^1$ which corresponds to segment 1 in Fig \ref{fig:demo_timeline} (top). Similarly, each state transition corresponds to an action and for each action, a set of cartesian DMPs are trained. Here the target object is the handle and the goal object is the whole drawer set.
\begin{table}[t]
\resizebox{\columnwidth}{!}{
\begin{threeparttable}[b]
\small
\centering
\caption{Success trials and timings for our 7 evaluation tasks}
\label{table:success_rate}
% \begin{tabular}{{p{1.5cm}|p{1.8cm}|p{1.8cm}|p{0.8cm}|p{0.8cm}}
\begin{tabular}{c|c|c|c|c}
\hline
\textbf{Tasks} & \textbf{Success Trials} & \textbf{Success Trials} & \textbf{Demo.} & \textbf{Prog.}\\
& \textbf{(Normal)} & \textbf{(Reactive)} & \textbf{Time} & \textbf{Time}\\
\hline
\textbf{P\&P} & 20/20 & 20/20 & 23.1s &7s\\
\textbf{Insert} & 16/20& 5/5& 23.5s& 6.7s\\
\textbf{Erasing} & 19/20& 5/5& 26.4s& 7s\\
\textbf{Drawer} & 18/20& 4/5& 26s & 8.5s\\
\textbf{Pouring} & 17/20& 5/5 & 24.5s& 8.6s\\
\textbf{Kitting\tnote{*}} & 19/20 & 5/5& 0s&0.5s\\
\textbf{Drawer P\&P}\tnote{*}& 19/20& 4/5& 0s& 0.5s\\
\hline
\end{tabular}
\begin{tablenotes}
    \item [*] Composite BT
\end{tablenotes}
\end{threeparttable}
}
\end{table}
The segmented state variables are converted into the tuple $\mathcal{T}$=$\{(Action^1$, $[e_2({\tt Near})])$, $(Action^2$, $[g_3({\tt Closed})$,$ e_3({\tt Near})])$, $(Action^3$,$[o_4({\tt Goal})$, $e_4({\tt Near})])$, $(Action^4$, $[g_5({\tt Open})$, $o_5({\tt Goal})])$, $(Action^5$, $[o_6({\tt Goal})$, $e_6({\tt !Near})])\}$ through the Algorithm \ref{alg:BTgen} function {\tt CondAbstraction}. $\mathcal{T}$ is then iteratively converted into a BT by chaining atomic BTs in the same sequence. Fig. \ref{fig:Overview} shows the generated BT for the \textit{drawer} task. Demonstration data, segmented state variables and generated BTs for all the 7 tasks are available at this \href{https://github.com/jainaayush2006/CoBT.git}{link}.

For performance evaluation 20 trials were conducted for each task, where target and goal object's start pose are changed within \textit{80cm}x\textit{40cm} area on the table between trials. An equal amount of reverse trials, where the opposite trajectory to the demonstration is needed, are performed except for \textit{drawer} and \textit{drawer P\&P} task due to the limitations in the robot's workspace. All trials are slowed at least 2 times using DMP time scaling to ensure safety. The number of successful trials in normal conditions for each task is shown in Table \ref{table:success_rate}. Notably, the \textit{P\&P} task achieved a 100\% success rate, while the \textit{insert} task exhibited the highest number of failures. During 20 attempts to execute the \textit{insert} task, there were 4 instances where the cube did not smoothly slide into the slot, primarily due to vision-related errors. These errors included issues such as failure to grasp the target object in the correct pose and inaccuracies in computing the goal pose. Indeed vision errors emerged as a dominant factor contributing to failures across all tasks. 2 \textit{drawer} task failures, 1  \textit{drawer P\&P} task failure and 1 \textit{pouring} task failure were attributed to robot joint limits. The remaining 2 \textit{pouring} task failures occurred during reverse trajectory trials. In these cases, the robot tried to move the cup to the opposite side of the goal object to initiate the pouring action resulting in collisions since the learned transformation $^{\mathcal{O}}\textbf{T}_e$ is not reversed in the reverse trajectory. In normal conditions with varying start poses, CoBT achieved over 91\% success rate and the average programming time from demonstration amounted to 7.5s. 

\subsection{Reactivity}\label{subsec:reactivity}
To evaluate the reactivity, each task was executed at least 5 times, with variations in object positions during executions. The \textit{P\&P} task's reactivity was evaluated over 20 trials as all manipulation tasks incorporate elements of the \textit{P\&P} action. For each sub-task within \textit{P\&P}, i.e., reaching the target object, grasping, moving toward the tray, and placing the object, we altered the object's position 5 times. Table \ref{table:success_rate} illustrates the number of successful trials during reactive conditions showing that CoBT-generated programs exhibit a high degree of reactivity, with nearly 100\% reactivity to environmental changes.
When an object, target or goal position is altered during execution, the effect conditions for that particular segment will not be successful. Therefore, BT will reattempt that particular segmented of action until all the effect conditions are met. 
The reactivity is due to the fact that the system not only generalizes on the motion level but on the task level. Due to state-action mapping during the segmentation process, the BT will execute the same action in the loop until the success state is achieved. 
Furthermore, even if the task is reset afterward, the BT will keep running in the loop until the task's goal conditions succeed. The 2 instances of failed reactive trials occurred while attempting to open the \textit{drawer}. These failures were attributed to the constraints imposed by the robot's joint limits.

\subsection{Modularity}\label{subsec:modularity} 
In this experiment, we show the modularity of our generated BT. A user-specified goal $\mathcal{G}$ generates a composite BT through algorithm \ref{alg:Goal_abs}. In the \textit{kitting} task, the system is shown three objects placed on a tray and for \textit{drawer P\&P} task, a cube inside the opened drawer is shown. Using algorithm \ref{alg:Goal_abs}, the system adapts goals of individually demonstrated and generated BTs and sequences them. For \textit{kitting} the target objects are the three black pieces and goal object is the tray, and for \textit{drawer P\&P} the target objects are the handle and cube and the goal object is the drawer set. The demonstration for all the target object are recorded individually.

To evaluate the modularity, \textit{kitting} and \textit{drawer P\&P} tasks are executed for the same 20 normal and 5 reactive times. The results in Table \ref{table:success_rate} show that each task had only one failure during normal trails and only \textit{drawer P\&P} task failed once in reactive trial. These tasks \textit{did not} require new demonstrations as they were adapted from previously existing demonstrations i.e. the \textit{kitting} task is simply three different \textit{P\&P} tasks sequenced and \textit{drawer P\&P} task is \textit{drawer} and \textit{P\&P} task sequenced.

\subsection{Pilot-study} \label{subsec:user_study}
Finally, a pilot study is conducted to verify the accessibility for novice subjects for \textit{P\&P} and \textit{drawer} tasks, involving N=10 participants (2 females, 8 males, aged 22-40). Out of 10 only 5 had prior experience with collaborative robots, and 0 with PbD systems. At the start, participants could familiarize themselves with hand-guiding mode and gripper controls. Subsequently, one demonstration for each task was recorded. They were also allowed to re-try their demonstrations if dissatisfied. Importantly, no additional PbD-specific instructions were provided. Further details about the step-by-step procedure of the study, participant surveys conducted and the ethics approval are available at this \href{https://github.com/jainaayush2006/CoBT.git}{link}.

\begin{figure}[!t]
\centering
\includegraphics[width=\columnwidth]{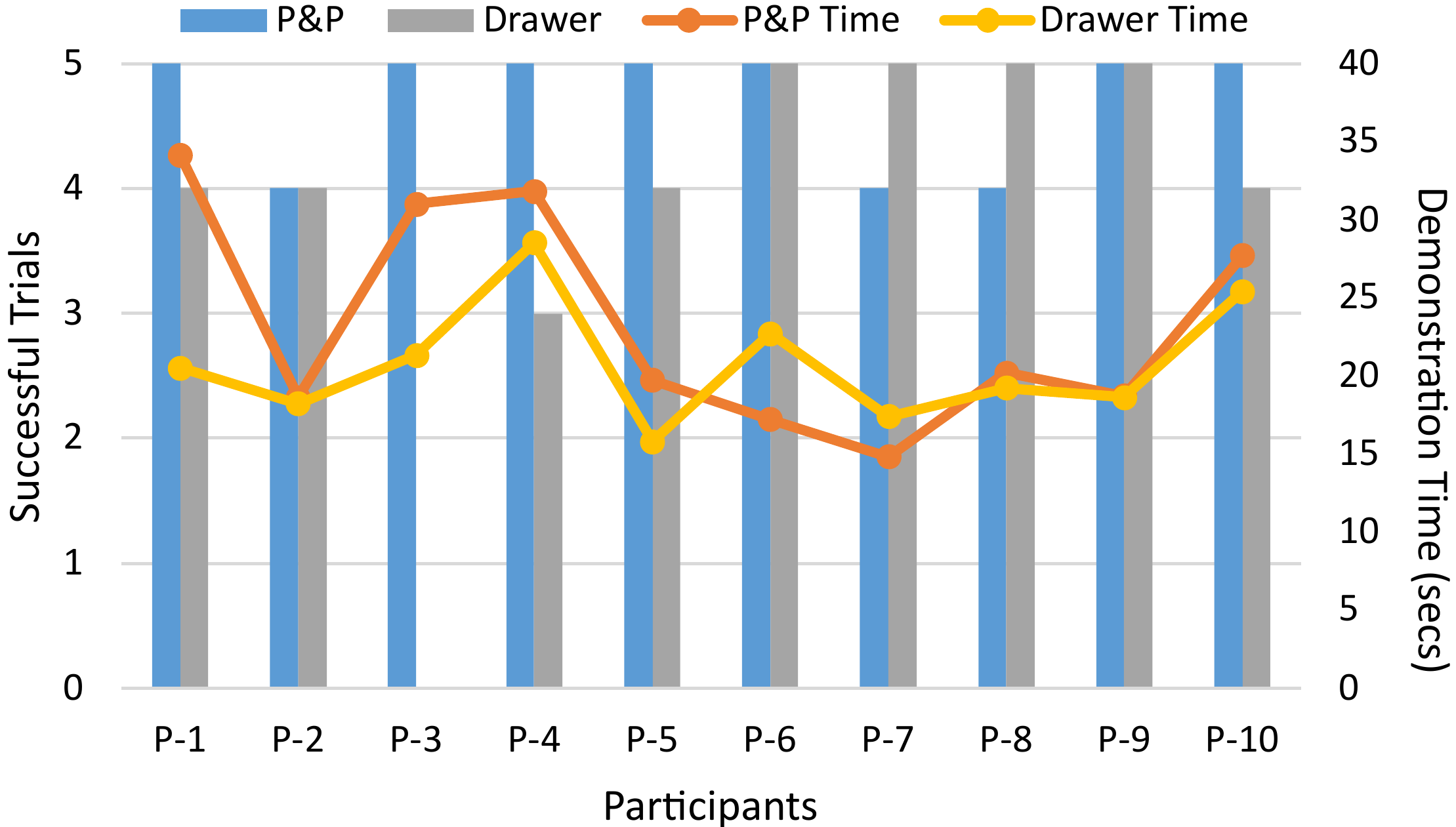}
\caption{Successful trials and demonstration time for each participant.}
\label{fig:User_study}
\end{figure}

Fig. \ref{fig:User_study} shows the number of successful trials out of 5 for each demonstration and demonstration times for each participant. Among 20 non-expert demonstrations, only one failed (P3, \textit{drawer} task) in programming a deployable BT due to conflicting effect conditions of consecutive actions and was excluded from analysis. In a single case (P4), the segmentation algorithm combined two actions (grasping the handle and opening the drawer) in the \textit{drawer} task, resulting in a BT that failed in 2 out of 5 trials but remained deployable with reduced reactivity. Minor variations in retracting the robot after object placement during the \textit{P\&P} task led to 2 failed trials. Notably, the two non-experts (P9 and P6) with a 100\% success rate lacked prior experience with collaborative robots. The remaining failures were attributed to the quality of demonstrated motion and the resulting DMPs derived from them. Overall, CoBT-generated programs from non-expert demonstrations achieved a 94\% success rate for the \textit{P\&P} task and approximately 87\% for the \textit{drawer} task. Finally, we conducted the raw NASA-TLX questionnaire\cite{hart1988development}, with scores ranging 1-7, to measure cognitive load and a Systems Usability Scale (SUS) questionnaire\cite{brooke1996sus} with scores ranging 1-5 which was further scaled between 1-100 for final score. On average, participants rated \textit{mental demand} at 2.1 and \textit{effort} at 2.7, while \textit{physical demand} was rated at 3.10. In general, participants found the system easy to use, with average SUS score of 92, and perceived a low requirement for prior knowledge, with average SUS score of 28. In summary, both quantitative and qualitative data confirm CoBT's ease of use.
\section{Conclusion and Future Work} \label{sec:conclusion}
In this paper, CoBT, a collaborative programming framework to generate BTs from one demonstration, is presented. CoBT generates deployable, reactive and modular programs in the form of DMP embedded BT and does not require any prior task or object specific knowledge. We have shown CoBTs capability to generalise the task, the reactivity to environmental changes and modularity on 7 evaluation tasks. Our system achieved $\approx$ 93\% success rate overall with an average of 7.5s programming time. We further presented a pilot study with non-expert participants (N=10) that demonstrates the system's ease of use. To the best of our knowledge, CoBT is the only framework capable of generating deployable BTs from a single demonstration. The current framework offers the potential for agile robot programming in industry. Nonetheless, there are certain limitations. CoBT's execution accuracy is limited to users demonstration precision just like any hand-guiding system. The DMP generalizes from a single demonstration, thereby inheriting associated limitations, such as the absence of stochasticity and challenges in generalizing reverse executions. Additionally, the generated BTs may benefit from post-processing to address conflicting constraints, such as with the non-expert demonstration, utilizing interactive human inputs, as suggested in \cite{iovino2022interactive}, a direction we intend to explore in future research.

% \clearpage
\bibliography{references}{}
\bibliographystyle{IEEEtran}

\end{document}